\UseRawInputEncoding
\documentclass[a4paper,10pt,oneside]{article}
\usepackage{amsmath,epsfig,times,url,hyperref,booktabs,graphicx}

\title{A Minimalist Brain--Computer Musical Interface for Real-Time Emotion-Driven Sonification: System Design and Preliminary Evaluation}

\date{}
\begin{document}
%\ninept
\maketitle

% IF ONE AUTHOR
\begin{center}
\author{Pablo A. Monroy-D'Croz$^{1}$, 
Rafael Ramirez-Melendez$^{2}$, 
Julian Cespedes-Guevara$^{3}$} \\
{$^{1}$ Universitat Pompeu Fabra, Barcelona, Spain \\
{\tt pablo@monroy.net}
\\[1ex]
$^{2}$ Universitat Pompeu Fabra, Barcelona, Spain\\
{\tt rafael.ramirez@upf.edu}
\\[1ex]
$^{3}$ Universidad Icesi, Cali, Colombia \\
{\tt jcespedes@icesi.edu.co}}
\end{center}

\begin{sloppy}

\begin{abstract}
This paper presents a minimalist Brain–Computer Musical Interface (BCMI) that functions as a real-time affective sonification system, translating prefrontal EEG activity into adaptive music. Emotional valence is estimated from frontal alpha asymmetry (AF7/AF8) and mapped to musical features such as mode, tempo, rhythmic density, and pitch register through a stochastic generative algorithm. The system integrates wireless EEG acquisition, real-time Python signal processing, and Ableton Live-based music generation synchronized via Lab Streaming Layer.
An experiment with 22 participants investigated whether intentional emotional self-induction (happy vs. sad) could modulate the BCMI neurofeedback signal. Linear mixed-effects analyses found no significant effects of target emotion or time, indicating that the frontal alpha asymmetry signal did not reliably distinguish instructed emotional states. Individual differences, including musical training and acting experience, explained more variance than the experimental manipulation, which accounted for only 0.40\% of total signal variance. These findings highlight the challenges of using frontal alpha asymmetry as a voluntary control signal for closed-loop emotion regulation and suggest methodological directions for future BCMI research.

\end{abstract}

\section{Introduction}

The translation of physiological signals into sound sonification has become a well established method for representing data streams that are otherwise inaccessible to conscious perception \cite{hermann2011sonification}. Within this broad domain, \textit{affective sonification} denotes the mapping of emotional or psychophysiological states onto auditory parameters, creating a perceptual representation of an individual's ongoing affective dynamics. When the sonified signal is musical in structure, such systems intersect with the field of \textit{musification}, where data drive aesthetically coherent musical output rather than abstract sound \cite{vickers2016aesthetics,supper2014sublime}.

Music has intrinsic emotional connotations. Empirical research in music psychology has established systematic relationships between acoustic features (tempo, mode, timbre, rhythmic density) and listener affective responses \cite{Juslin2019,Koelsch2014}. This makes music a particularly compelling medium for affective sonification: changes in the underlying physiological signal can be mapped to musical transformations that listeners are predisposed to interpret in emotional terms. When such mapping operates in a closed loop, where the listener's own physiological state drives the music and the music in turn influences that state, the resulting interaction can be understood as a form of \textit{neurofeedback}, a process with potential applications in emotional self-regulation, stress management, and therapeutic intervention \cite{Sitaram2017,Ramirez2015}.

Brain--Computer Musical Interfaces (BCMIs) extend traditional BCIs by mapping neural signals to musical processes \cite{miranda2014guide,williams2018bci}. In affective BCMIs, neural correlates of emotion, such as frontal alpha asymmetry, serve as control signals for real-time music generation, creating a bidirectional coupling between brain activity and auditory feedback \cite{ehrlich2019closedloop,daly2018affective}. However, many existing BCMI systems rely on dense EEG electrode arrays, complex laboratory setups, or computationally expensive machine learning pipelines that limit their accessibility, portability, and reproducibility outside specialized research environments \cite{Lin2010}. There remains a need for minimalist architectures that can support real-time affective interaction using lightweight, low-cost sensing hardware.

This paper investigates whether a minimalist EEG configuration, specifically two prefrontal electrodes (AF7/AF8), can support real-time emotion-driven music generation within a closed-loop BCMI framework. The system estimates emotional valence using the Asymmetric Frontal Activity Hypothesis (AFAH), a well-established neuroscientific model linking differential prefrontal alpha-band activity to approach-withdrawal motivation \cite{Davidson1990,Allen2018}. This estimate is then mapped to musical parameters via a stochastic rule-based generative engine adapted from Ehrlich et al.\ \cite{ehrlich2019closedloop}.

The contributions of this work are:

\begin{enumerate}
    \item A fully specified, reproducible BCMI architecture for real-time affective sonification using a two-channel prefrontal EEG configuration, wireless biosignal acquisition, and a parameterizable generative music engine integrated with a commercial digital audio workstation.
    
    \item A formal description of the emotion-to-music mapping, detailing the transformation from continuous valence estimates to harmonic mode, tempo, rhythmic density, pitch register, and loudness, including the probabilistic rules governing musical output.
    
    \item A preliminary experimental evaluation ($N = 22$) of the system's capacity to reflect voluntary emotional self-induction, employing linear mixed-effects modeling to assess signal dynamics, quantify variance components, and characterize individual differences--with the aim of establishing an empirical baseline for future methodological refinements.
\end{enumerate}

Importantly, this paper reports the experimental outcomes as they were observed, including null findings where they occurred. We present these results not as a demonstration of successful emotion decoding (the data do not support such a claim), but as a transparent empirical characterization of the limitations of the AFAH-driven approach in a volitional, closed-loop context and as a methodological foundation for the next generation of affective BCMI systems.

\section{Related Work}

\subsection{Theoretical Models of Emotion}

Two dominant frameworks structure the scientific study of emotion. \textit{Categorical models} posit a limited set of discrete, biologically basic emotions--such as happiness, sadness, anger, and fear--that are universally expressed and recognized \cite{Ekman01051992}. While influential in psychological research, categorical models provide limited traction for computational systems that require continuous control signals.

\textit{Dimensional models} conceptualize emotions as positions within a continuous multidimensional space. The circumplex model of affect \cite{Russell1980} organizes emotional experience along two orthogonal axes: \textit{valence} (pleasantness--unpleasantness) and \textit{arousal} (activation--deactivation). This representation has become standard in affective computing because it supports gradual transitions, avoids linguistic ambiguity, and enables systematic mapping between emotional parameters and stimulus features \cite{Wu2023}. The present work adopts a unidimensional projection of the circumplex, mapping the diagonal axis from low-valence/low-arousal to high-valence/high-arousal onto a continuous scalar score [0,1].

\subsection{Music and Emotion}

Music influences emotional states through multiple psychological mechanisms, including brainstem reflexes to acoustic features, emotional contagion via perceived expression, episodic memory associations, and violations or fulfillments of learned syntactic expectations \cite{Juslin2019}. These mechanisms operate in parallel and interactively, explaining why the same musical excerpt can evoke different emotional responses across individuals and contexts.

Empirically, specific musical features map systematically onto affective dimensions. Fast tempi and high loudness generally increase perceived arousal, while major mode and consonant harmony are associated with positive valence \cite{Juslin2010,Gagnon2003}. Timbral brightness, pitch height, and articulation further modulate affective interpretation. These relationships--summarized in Table~\ref{tab:music_features_emotion}--provide the empirical foundation for the emotion-to-music mapping implemented in the present BCMI.

\begin{table}[htbp]
\centering
%\footnotesize

\caption{Common musical features and their typical associations with emotional valence and arousal. Adapted from \cite{juslin2010expression,Juslin2019}.}
\label{tab:music_features_emotion}
\begin{tabular}{@{}p{3.6cm}p{4.0cm}p{3.6cm}@{}}
\toprule
\textbf{Musical Feature} & \textbf{Valence Association} & \textbf{Arousal Association} \\
\midrule
Fast tempo & Neutral or slightly positive & High \\
Slow tempo & Neutral or slightly negative & Low \\
Major mode & Positive & Neutral or moderate \\
Minor mode & Negative & Neutral or moderate \\
Consonant harmony & Positive & Low to moderate \\
Dissonant harmony & Negative & High \\
High loudness & Context-dependent & High \\
Low loudness & Context-dependent & Low \\
High average pitch & Positive & Moderate to high \\
Low average pitch & Negative or powerful & Low to moderate \\
\bottomrule
\end{tabular}
\end{table}

An essential distinction in music-emotion research concerns \textit{perceived} versus \textit{induced} emotion \cite{Juslin2010}. Perceived emotion refers to the emotional quality recognized in the music; induced emotion refers to the actual affective state experienced by the listener. These can diverge: a listener may perceive a piece as sorrowful yet feel comforted or moved. Affective BCMI systems that aim to support emotional regulation must ultimately target induced states; however, the present work focuses on whether the system's output signal reflects voluntary shifts in neural activity associated with emotional intention, without making claims about whether the music successfully induced those states.

\subsection{EEG-Based Emotion Detection and Frontal Alpha Asymmetry}

Electroencephalography (EEG) provides non-invasive access to neural dynamics with millisecond temporal resolution, making it well-suited for tracking rapidly evolving affective processes \cite{Cui2022}. Among the most extensively studied EEG correlates of emotion is \textit{frontal alpha asymmetry} (FAA).

According to the Asymmetric Frontal Activity Hypothesis (AFAH), hemispheric lateralization in prefrontal alpha-band (8--13 Hz) activity reflects motivational direction underlying emotional experience \cite{Davidson1990}. Because alpha power is inversely related to cortical activation \cite{Pfurtscheller1999}, relatively lower left-frontal alpha (indicating greater left prefrontal activation) is associated with approach-related motivation and positive affect, whereas relatively lower right-frontal alpha is associated with withdrawal-related motivation and negative affect \cite{Allen2018,Reznik2018}. FAA is typically operationalized as the difference in alpha power between homologous left and right frontal electrodes (e.g., F3/F4 or AF7/AF8).

Despite its theoretical grounding, FAA exhibits important psychometric limitations. Test-retest reliability is moderate within a session but only fair across sessions, and estimates are sensitive to transient mood, arousal, and task engagement \cite{KollerSchlaud2020}. Resting FAA often shows weak or non-significant correlations with self-reported affective traits, and meta-analyses have reported very small effect sizes for FAA as a diagnostic marker of depression or anxiety \cite{VanDerVinne2017,Luo2025}. FAA effects are most robust under conditions that maximize motivational salience--emotionally evocative films, social interactions, or incentive-driven tasks--rather than passive resting states \cite{Mennella2017,Sabu2022}.

These findings carry direct implications for the present work: while FAA provides a theoretically grounded and computationally efficient method for real-time affect estimation, its application as a control signal for volitional emotion self-induction in a closed-loop BCMI may be constrained by the very limitations identified in the broader literature.

\subsection{Brain--Computer Musical Interfaces and Affective Sonification}

Brain--Computer Musical Interfaces map neural signals to musical processes, enabling users to influence sound generation through brain activity \cite{miranda2014guide}. Recent work has explored BCMIs in therapeutic neurofeedback \cite{Ramirez2015}, interactive performance \cite{hopkins2023stringesthesia}, and emotional regulation \cite{Hildt2021}.

Of particular relevance is the closed-loop music-based BCI developed by Ehrlich et al.\ \cite{ehrlich2019closedloop}, which demonstrated that a parameterizable music generation algorithm could serve as both calibration stimulus and real-time feedback, with emotional parameters inferred from 14-channel EEG driving affective musical output in a continuous loop. The present work adapts Ehrlich et al.'s music generation framework to a minimalist two-channel configuration, evaluating whether their closed-loop architecture scales down to portable, low-cost hardware.

The intersection of sonification and affect has been explored through the concepts of \textit{musification}--the translation of data into aesthetically coherent musical structures--and the aesthetic turn in sonification research, which emphasizes the listener's experiential engagement with the resulting sound \cite{vickers2016aesthetics,supper2014sublime}. The present BCMI can be understood as a musification system in which the data source is the user's own neural activity, and the aesthetic experience of the generated music is intended to support, rather than merely represent, the underlying affective process. We adopt Supper's \cite{supper2014sublime} framing of sonification as a hermeneutic encounter: the listener does not merely decode the mapping but actively interprets the musical output within a personal and cultural context. This framing acknowledges that the emotional meaning of the generated music is co-constructed by the listener's background, expectations, and current state--a consideration that complicates any simple stimulus--response model of affective sonification.

\subsection{Aesthetic Positioning and Cultural Tradition}

The generative music system described in this paper operates within the Western tonal tradition, employing seven diatonic modes (Lydian through Locrian), four-bar I--IV--V--I chord progressions, and chamber-orchestra timbres (piano, cello, double bass). The emotion-to-music mappings--major modes for positive valence, fast tempi for high arousal--likewise reflect Western musical conventions \cite{Gagnon2003}. We acknowledge that these associations are culturally specific and may not generalize to listeners enculturated in non-Western tonal systems. This limitation is particularly salient given that the system is intended for human interaction: the perceived emotional character of the generated music depends on the listener's internalized musical grammar. Future iterations should explore cross-cultural mapping strategies or incorporate user-specific calibration of affective musical parameters.

\section{System Architecture}

The BCMI integrates four primary subsystems within a closed-loop architecture (Figure~\ref{fig:architecture}): (1) wireless two-channel EEG acquisition, (2) real-time signal processing and affect estimation, (3) stochastic rule-based music generation, and (4) a digital audio workstation (DAW) for sound synthesis and auditory instruction playback (Ableton Live). All software components were implemented in Python 3.9 as a multithreaded application, with the Lab Streaming Layer (LSL) protocol \cite{kothe2025lsl} providing synchronized multimodal data recording in XDF format via LabRecorder.

\begin{figure}[htbp]
    \centering
    \includegraphics[width=0.95\linewidth]{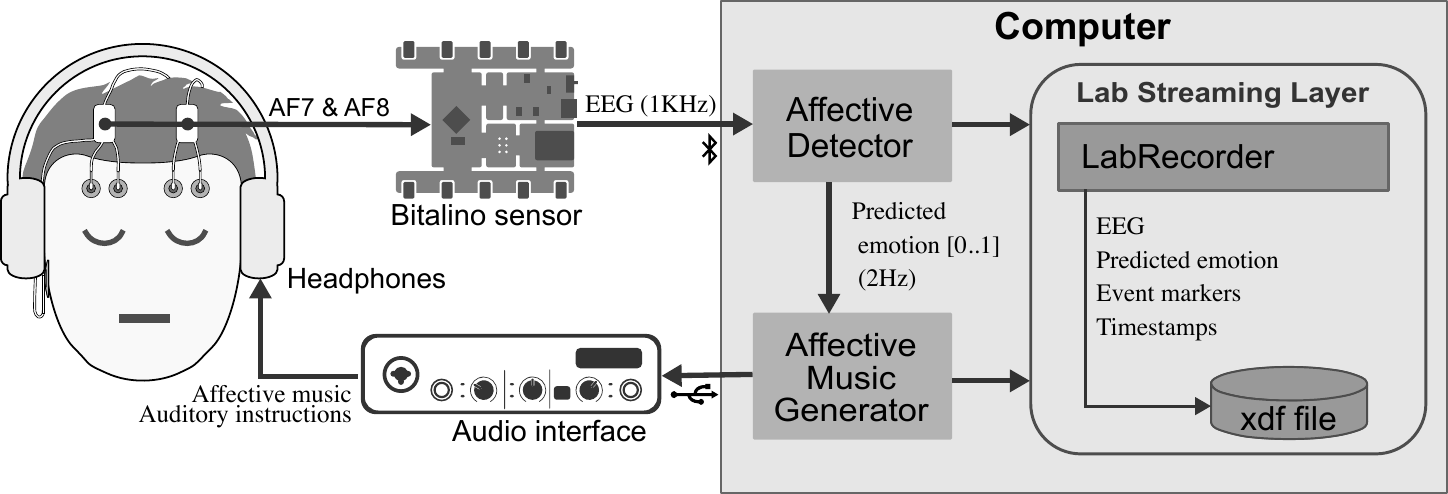}
    \caption{System architecture of the BCMI. EEG signals from AF7/AF8 are acquired wirelessly (BITalino), processed in real time to extract frontal alpha asymmetry and compute a continuous emotion estimate, and transmitted  the affective music generator, which produces MIDI messages driving virtual instruments in a DAW. Auditory feedback is delivered through closed-back headphones, closing the neurofeedback loop. Experimental data is recorded via LSL}
    \label{fig:architecture}
\end{figure}

\subsection{EEG Acquisition Hardware}

Neural signals were acquired using a BITalino (r)evolution Plugged Kit BT sensor (PLUX Wireless Biosignals S.A., Portugal) \cite{da2014bitalino}, featuring Bluetooth connectivity and five reusable Ag/AgCl electrodes. The recording configuration employed a two-channel bipolar EEG montage with differential amplification, using electrodes positioned at international 10--10 system locations AF7 (left anterior frontal) and AF8 (right anterior frontal), with a reference electrode at A2 (right mastoid). Each bipolar channel used paired electrodes with 25~mm center-to-center spacing. Conductive paste (Ten20; Weaver and Company, USA) was applied before electrode placement, and an elastic cloth headband secured the electrodes, maintaining impedances below 5~k$\Omega$. EEG signals were sampled at 1000~Hz and transmitted wirelessly to the central processing unit.

\subsection{Software Architecture and Signal Processing}

The BCMI software was structured as a multithreaded Python application. The primary execution thread managed the experimental protocol and invoked the affective music generator at predefined intervals. A parallel secondary thread performed continuous emotion detection, running the affective detector module on a sliding window of 4~seconds with 87.5\% overlap, yielding an updated emotion estimate every 0.5~seconds.

Power spectral density (PSD) was computed for the AF7 and AF8 channels using Welch's method \cite{Welch1967}. Within each 4-second window, alpha-band (8--13~Hz) and beta-band (14--29~Hz) power were extracted. Arousal and valence indices were derived following \cite{Ramirez2020}:

\begin{equation}
\textit{arousal (A)} = \frac{\beta_{\mathrm{AF7}} + \beta_{\mathrm{AF8}}}{\alpha_{\mathrm{AF7}} + \alpha_{\mathrm{AF8}}}
\end{equation}

\begin{equation}
\textit{valence (V)} = \frac{\alpha_{\mathrm{AF8}}}{\beta_{\mathrm{AF8}}} - \frac{\alpha_{\mathrm{AF7}}}{\beta_{\mathrm{AF7}}}
\end{equation}

A two-stage normalization procedure was applied to stabilize the estimates. First, a local min-max normalization was computed over a 20-second moving window. Second, a long-term adaptive normalization was applied, recursively updating global minima and maxima to maintain the signal within [0,1]. The final predicted emotion $E_t$ was computed as the arithmetic mean of the normalized arousal and valence values. A 10-second moving average of $E_t$ was additionally computed to provide the smoothed valence estimate $\bar{V}_t$ that determined harmonic mode selection at the onset of each four-bar musical progression, ensuring temporal coherence between the affective input and the generative musical structure.

\subsection{Affective Music Generation}

The affective music generator implements a stochastic rule-based architecture adapted from \cite{ehrlich2019closedloop}, comprising a generative harmonic sequencer and a configured digital audio workstation (DAW) environment. The system constructs four-bar chord progressions (I--IV--V--I) cyclically, operating at one-eighth-note temporal resolution, and transmits MIDI note events through a preconfigured output port using the Mido Python library.

\subsubsection{Harmonic Mode Selection}

Seven diatonic modes were defined as ordered four-bar chord progressions, organized from most positive to most negative emotional valence following the \textit{Kirchentonleitern} ordering: Lydian (mode 0), Ionian (mode 1), Mixolydian (mode 2), Dorian (mode 3), Aeolian (mode 4), Phrygian (mode 5), and Locrian (mode 6). The active modal index $m(t)$ was selected at the beginning of each four-bar cycle by quantizing the smoothed valence estimate:

\begin{equation}
m(t) = 6 - \operatorname{round}\!\big(6 \cdot \bar{V}_t\big)
\end{equation}

such that $\bar{V}_t \rightarrow 1$ (highest valence) yields Lydian mode, and $\bar{V}_t \rightarrow 0$ (lowest valence) yields Locrian mode. Each mode $m$ was realized as a sequence of four chords drawn from a predefined chord matrix, with each chord comprising four MIDI note numbers corresponding to piano accompaniment, cello doubling, bass, and melodic voices.

\subsubsection{Affect-Dependent Musical Parameters}

The predicted emotion components were mapped to musical control variables as follows:

\begin{itemize}
    \item \textbf{Tempo:} The inter-tick interval $\Delta t$ was computed as $\Delta t = 0.3 - 0.15 \cdot A(t)$, where $A(t)$ is the arousal estimate. This produces tempi ranging from approximately 100~BPM (low arousal) to 200~BPM (high arousal).
    \item \textbf{Rhythmic density:} The roughness parameter $r(t) = 1 - A(t)$ determined the probability of note activation at each tick. For each tick $k$ and voice $j$, a note was triggered if $U_k^{(j)} \geq r(t)$, where $U_k^{(j)} \sim \mathcal{U}(0,1)$. Expected activations per tick were thus $2 \cdot A(t)$, ranging from sparse textures at low arousal to dense polyphonic activity at high arousal.
    \item \textbf{Loudness:} MIDI velocity was sampled uniformly: $l \sim \operatorname{UnifInt}\{50, \dots, \operatorname{round}(40 \cdot A(t)) + 60\}$, producing a dynamic range from piano to forte as arousal increased.
    \item \textbf{Pitch register:} Voicing $\phi(t) = V(t)$ controlled octave transpositions. Accompaniment pitches received stochastic octave displacements $B_i \in \{-1, 0, 1\}$ biased upward when $V(t) > 0.5$ and downward otherwise. Bass register was set to the root pitch minus 12 semitones for high-voicing conditions and minus 24 semitones for low-voicing conditions.
\end{itemize}

\subsubsection{Digital Audio Workstation Integration}

Ableton Live 11 served as the DAW, hosting virtual instruments that emulated a small chamber orchestra: a grand piano performing both accompaniment and melodic material (MIDI channels 1--2), a double bass providing the bass line (channel 3), and a cello section doubling the accompaniment (channel 4). Pre-recorded auditory instructions were generated using the Narakeet text-to-speech platform (voice model ``Lola,'' Latin American Spanish) and embedded within the DAW session, triggered by MIDI continuous controller messages mapped to experimental protocol events. Audio output was delivered through a Behringer U-PHORIA UMC22 USB audio interface to Sony MDR-V6 closed-back monitoring headphones, providing passive ambient noise isolation.

\subsection{Closed-Loop Operation and Data Logging}

The integration of EEG sensing, affect estimation, and generative music synthesis established a closed-loop interaction: neural signals influenced music generation, and the resulting music could in turn influence the participant's neural and emotional state. All data streams--raw EEG (1000~Hz), predicted emotion and smoothed valence (2~Hz), MIDI event markers, and experimental event markers--were synchronized and recorded in a unified XDF file via LSL, enabling complete post-hoc reconstruction of each experimental session. The computational platform was a MacBook Pro M1 (2020) with 16~GB unified memory, running macOS Big Sur 11.7.

\section{Experiment}

To evaluate whether the BCMI system could reflect volitional emotional self-induction, an experimental study was conducted in which participants attempted to induce happy and sad emotional states while receiving real-time music generated from their own EEG activity. The study employed a within-subjects, repeated-measures design. The protocol was approved by the Human Ethics Committee of Universidad Icesi under Act \#578.

\subsection{Participants}

Twenty-two healthy volunteers (8 female; mean age = 33.8 years, SD = 12.6, range: 18--59) were recruited from the student and staff population of Universidad Icesi. All were native Spanish speakers with self-reported normal hearing and no history of neurological or psychiatric disorders. Twelve participants had formal musical training, six had formal acting training, and three reported taking psychoactive medication. Written informed consent was obtained from all participants.

\subsection{Experimental Protocol}

Each participant completed 16 trials of a target emotion self-induction task (8 happy, 8 sad), presented in a semi-random order constrained so that no more than two consecutive trials shared the same condition. Each trial comprised three phases (Figure~\ref{fig:protocol}):

\begin{enumerate}
    \item \textbf{Silent baseline (14~s):} Participants rested in silence to allow emotional resetting from the previous trial.
    \item \textbf{Auditory instruction ($\sim$4~s):} A pre-recorded verbal cue (``happy'' or ``sad,'' in Spanish) indicated the target emotion.
    \item \textbf{Music feedback (30~s):} The BCMI generated music in real time, with the AFAH-derived emotion estimate driving the affective music generator as described in Section~3. Participants were instructed to self-induce the target emotion using personal strategies (e.g., autobiographical recall, mental imagery) while attending to the music.
\end{enumerate}

\begin{figure}[htbp]
    \centering
    \includegraphics[width=0.95\linewidth]{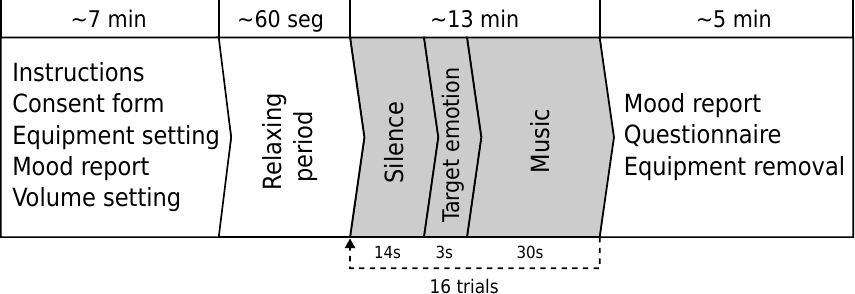}
    \caption{Structure of a single experimental trial. Each trial began with a 14-second silent baseline, followed by a 4-second auditory instruction indicating the target emotion (happy or sad), and concluded with 30 seconds of real-time BCMI-generated music driven by the participant's EEG activity.}
    \label{fig:protocol}
\end{figure}

Prior to the experimental trials, participants completed a 60-second eyes-closed resting-state calibration block to initialize the system's normalization parameters. Baseline mood was assessed using the International Positive and Negative Affect Schedule--Revised (IPANAS-R) \cite{ruiz2020panas}. After the final trial, participants completed a post-experiment questionnaire assessing demographic characteristics, musical and acting training, emotion induction strategies, and subjective perceptions of the music's influence on their emotional state.

\subsection{Data Analysis}

Linear mixed-effects models (LMMs) were employed to analyze the hierarchical, repeated-measures data structure \cite{Yu2021,Koerner2017}. All analyses were conducted in R (version 2023.09.1) using the \textit{lme4} \cite{lme4}, \textit{lmerTest} \cite{lmerTest}, and \textit{clubSandwich} \cite{clubSandwich} packages.

\subsubsection{Model Specification}

The primary dependent variable was the continuous \textit{predicted\_emotion} signal (range [0,1], sampled at 1~Hz), representing the system's moment-to-moment estimate of emotional valence along the sad-to-happy diagonal of the circumplex model.

The random-effects structure was selected through sequential likelihood ratio tests comparing four nested models: (1) by-participant random intercept; (2) adding by-trial random intercept nested within participant; (3) adding by-participant random slope for time; (4) adding by-target-emotion random intercept within participant. The final random-effects structure included random intercepts for participant and trial, and a by-participant random slope for time.

The Base Model tested the primary experimental hypotheses:
\begingroup
\[
\begin{aligned}
\textit{predicted\_emotion} \sim\;&
\textit{target\_emotion} \times \textit{time} \\
&+ (1 \mid \textit{participant}) \\
&+ (0 + \textit{time} \mid \textit{participant})\\
&+ (1 \mid \textit{participant:trial})
\end{aligned}
\]
\endgroup

A Full Model additionally incorporated participant-level covariates: musician status, acting training, medication use, prior exposure to the BCMI paradigm, baseline positive affect (PANAS), age, and sex. Robust standard errors (cluster-robust variance estimator) were used for all inferences. Variance decomposition was performed using Nakagawa and Schielzeth's \cite{nakagawa2013general} $R^2$ statistics.

\section{Results}

\subsection{Primary Experimental Effects}

The Base Model analysis revealed no significant experimental effects. The main effect of target emotion was not significant ($\beta = -0.023$, 95\% CI $[-0.051, 0.005]$, $d = -0.12$, $p = .124$), indicating no reliable baseline difference in the neurofeedback signal between happy and sad trials. Critically, the target emotion × time interaction was not significant ($\beta = 0.000$, 95\% CI $[-0.002, 0.002]$, $d = 0.00$, $p = .666$), demonstrating that the linear trajectory of the predicted emotion signal did not diverge between conditions over the 30-second induction period. The main effect of time was also non-significant ($\beta = -0.001$, 95\% CI $[-0.003, 0.001]$, $d = -0.01$, $p = .168$).

Figure~\ref{fig:predicted_trajectories} displays the model-predicted trajectories by target emotion. The overlapping trajectories and wide confidence intervals illustrate the absence of systematic differentiation between conditions.

\begin{figure}[htbp]
    \centering
    \includegraphics[width=0.85\linewidth]{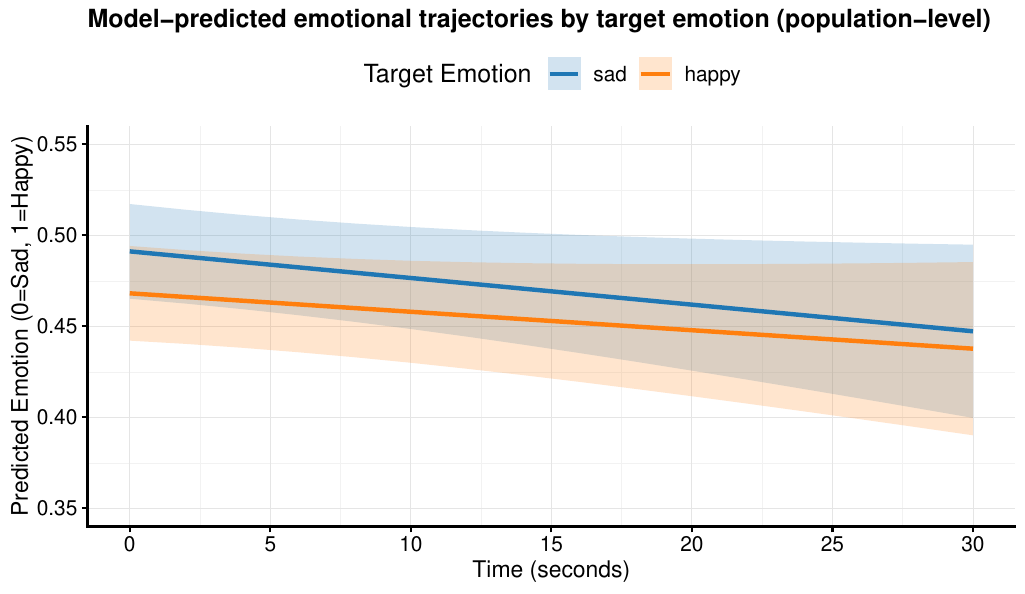}
    \caption{Model-predicted emotional trajectories by target emotion (Base Model). Shaded bands represent 95\% confidence intervals. The overlapping trajectories indicate no significant divergence between happy and sad conditions over the 30-second induction period.}
    \label{fig:predicted_trajectories}
\end{figure}

\subsection{Variance Decomposition}

The Full Model's variance decomposition (Table~\ref{tab:variance}) revealed that the experimental manipulation accounted for a negligible proportion of total signal variance (0.40\%). Participant-level covariates collectively explained 3.55\%, while the random-effects structure captured 19.60\%--dominated by trial-to-trial fluctuations (16.98\%). The residual, unexplained variance constituted 76.45\% of the total. The marginal $R^2$ (fixed effects only) was 0.037; the conditional $R^2$ (fixed + random effects) was 0.281.

\begin{table}[htbp]
\centering
\caption{Variance components of the Full Model. Experimental factors (target emotion × time) accounted for 0.40\% of total variance; residual and trial-level variability dominated.}
\label{tab:variance}
\begin{tabular}{@{}lrr@{}}
\toprule
\textbf{Component} & \textbf{Variance} & \textbf{\% Total} \\
\midrule
Experimental factors & 0.0002 & 0.40\% \\
Participant covariates & 0.0017 & 3.55\% \\
Random effects (total) & 0.0092 & 19.60\% \\
\quad Between-participant & 0.0012 & 2.60\% \\
\quad Trial-level & 0.0080 & 16.98\% \\
\quad Time slopes & 0.00001 & 0.02\% \\
Residual & 0.0360 & 76.45\% \\
\midrule
Total & 0.0471 & 100\% \\
\bottomrule
\end{tabular}
\end{table}

\subsection{Individual Differences}

Two covariates emerged as significant predictors. Musician status was positively associated with the neurofeedback signal ($\beta = 0.072$, 95\% CI $[0.032, 0.113]$, $d = 0.38$, $p = .005$), indicating that participants with formal musical training exhibited a higher overall predicted emotion signal. Acting training was negatively associated ($\beta = -0.046$, 95\% CI $[-0.079, -0.013]$, $d = -0.24$, $p = .026$). No other covariates reached significance.

\subsection{Qualitative Observations}

Post-experiment questionnaire responses indicated that participants employed diverse strategies for emotional self-induction, including autobiographical recall (e.g., memories of personal loss or achievement), mental imagery (e.g., visualizing happy or sad scenarios), and, in one case, meditative state induction. A subset of participants reported that the generated music ``matched'' their intended emotion at various points during the trials, though such reports were sporadic and not systematically related to the quantitative signal.

\section{Discussion}

The present study evaluated whether a minimalist, AFAH-driven BCMI could reflect intentional emotional self-induction in a real-time closed-loop configuration. The results provide a clear empirical answer: under the conditions tested, the system's neurofeedback signal did not differentiate between instructed happy and sad emotional states, either at baseline or in their temporal trajectories. This null finding warrants careful interpretation, as it does not necessarily indicate that participants failed to self-induce emotions--they may well have done so--but rather that the AFAH-derived signal, as operationalized in this two-channel configuration, lacked the sensitivity to detect such modulation.

\subsection{Interpreting the Null Findings}

Several factors may account for the absence of systematic experimental effects. First, FAA has been most reliably demonstrated under conditions of high motivational salience--emotionally evocative films, social interactions, or incentive-driven tasks \cite{Sabu2022,Mennella2017}. The present paradigm required participants to generate emotions volitionally through internal strategies, which may not engage the approach-withdrawal motivational systems that FAA primarily indexes, or may do so with substantially lower intensity than exogenous emotional stimuli.

Second, music processing itself engages distributed neural systems--including auditory, reward, and prefrontal networks--that may introduce neural activity unrelated to the target valence dimension \cite{Koelsch2014}. The concurrent demands of emotion regulation and musical appraisal may create cognitive competition for shared resources. Drawing on Ochsner and Gross's \cite{Ochsner2014} W-PVA (World--Perception--Valuation--Action) valuation framework, the experimental task can be understood as imposing three concurrent processes during the music-feedback phase: (4.1) internally directed emotion regulation, (4.2) basic auditory perception of the music, and (4.3) higher-order musical appraisal. When the cognitive load associated with (4.1) and (4.3) is high, their competition for controlled processing resources may degrade the neural signal available for valence estimation. This interpretation is consistent with evidence that the feedback stimulus itself can obscure the neural patterns that neurofeedback systems seek to detect \cite{Takabatake2021}.

Third, the limited spatial resolution of the two-channel montage, while justified by portability considerations, may be insufficient to capture the distributed network dynamics of emotion regulation. Volitional emotion generation recruits prefrontal, limbic, and parietal regions that a sparse frontal configuration cannot fully resolve \cite{Zotev2016AmygdalaFAA}.

\subsection{The Problem of Confounded Feedback}

A fundamental interpretive challenge in the present design is the inseparability of forward mapping (EEG$\rightarrow$music) and feedback effects (music$\rightarrow$EEG). The closed-loop architecture, while ecologically valid as a model of neurofeedback interaction, conflates two distinct hypotheses: whether EEG signals can drive emotionally congruent music, and whether that music in turn modulates the listener's emotional state. Without a no-music control condition or a condition where music is generated independently of the participant's EEG, it is impossible to attribute any observed signal variation to the participant's volitional effort versus their reaction to the music. Future studies should incorporate yoked control designs, in which participants receive music generated from another participant's EEG, to disambiguate these components.

\subsection{Individual Differences: Musicians and Actors}

The significant effects of musical and acting training, while modest in magnitude ($d = 0.38$ and $d = -0.24$, respectively), merit attention. Musical training strengthens auditory-motor coupling and promotes co-activation of perceptual and motor networks during auditory processing \cite{Bangert2006,Zatorre2007}. Such adaptations may enhance musicians' engagement with auditory feedback, amplifying the neurofeedback signal irrespective of target emotion. Acting training, by contrast, often emphasizes embodied, imagery-based, and interoceptive strategies for emotion generation that may recruit neural systems not directly reflected in frontal alpha asymmetry \cite{Brown2019Acting,SokolHessner2022}. These findings suggest that user background may substantively modulate BCMI output, reinforcing the need for personalized calibration strategies in future affective neurofeedback systems.

\subsection{Limitations}

Beyond the confounded feedback discussed above, several additional limitations should be noted. The 14-second inter-trial interval may be insufficient for complete emotional recovery; prior work suggests that recovery from musically or visually induced emotional states can require 2 minutes or more \cite{Ribeiro2019,gomez2009valence}. The repeated emotional transitions across 16 trials likely imposed cumulative cognitive demands, and the dominance of trial-level variance (16.98\%) is consistent with fatigue-related fluctuations. The study also lacked a direct measure of emotional induction success (e.g., continuous self-report), making it impossible to verify whether participants actually achieved the target states.

\subsection{Implications for Affective Sonification Design}

The present findings carry implications for the design of affective sonification systems. First, they underscore the gap between theoretical models validated in passive paradigms and their translation to active, closed-loop interaction. Sonification designers should anticipate that mappings that function reliably under controlled laboratory conditions may exhibit substantially degraded performance when the user is simultaneously engaged in regulatory tasks. Second, the high proportion of unexplained variance (76\%) highlights the inherently noisy nature of single-trial EEG-based affect estimation and suggests that robust affective sonification may require multimodal physiological integration--for example, combining EEG with electrodermal activity or heart rate variability--to improve signal stability.

\section{Conclusion}

This paper presented the design, implementation, and preliminary evaluation of a minimalist Brain--Computer Musical Interface for real-time affective sonification. The system integrates two-channel prefrontal EEG, AFAH-based affect estimation, and stochastic rule-based music generation within a closed-loop architecture that is fully specified and reproducible.

The experimental evaluation yielded null findings with respect to the system's capacity to reflect volitional emotional self-induction: linear mixed-effects models revealed no significant differentiation between happy and sad target conditions, and the experimental manipulation accounted for less than 0.5\% of total signal variance. Rather than interpreting these results as a failure of the BCMI concept broadly, we view them as establishing an empirical baseline that clarifies the limitations of the AFAH-driven approach in a volitional, closed-loop context--limitations that are consistent with the broader psychometric literature on frontal alpha asymmetry. The significant individual-difference effects (musician status, acting training) additionally highlight the importance of user-specific factors in BCMI design.

These findings motivate several directions for future work. First, the integration of complementary physiological measures--such as electrodermal activity and heart rate variability--may improve emotion estimation robustness without sacrificing the portability that distinguishes the minimalist approach. Second, machine learning methods trained on individualized calibration data may capture more complex neural patterns than a fixed, population-level asymmetry model. Third, the systematic investigation of musical parameters and their relationship to evolving neural signals requires fine-grained logging of generative features at each time step, enabling lagged analyses of closed-loop dynamics. Fourth, experimental designs incorporating yoked control conditions are essential to disambiguate forward and feedback mapping effects. Finally, perceptual validation studies are needed to assess whether listeners--both the users themselves and independent raters--can reliably interpret the emotional character of the generated music.

The BCMI platform described here, including its software implementation and complete experimental dataset, is publicly available  to facilitate replication and extension by the affective computing communities.\footnote{https://github.com/pamonroy/aBCMI} By transparently reporting both the system's architecture and its current empirical limitations, we hope to contribute a useful methodological foundation for the next generation of affective brain--computer musical interfaces.

\section{Acknowledgment}
\label{sec:ack}

This work was partially supported by Universitat Pompeu Fabra and Universidad Icesi. We thank the study participants for their time and engagement.

\bibliographystyle{IEEEtran}
\bibliography{references}

@article{Allen2018,
  author    = {Allen, John J. B. and Keune, Patrick M. and Sch{\"o}nenberg, Michael and Nusslock, Robin},
  title     = {Frontal {EEG} alpha asymmetry and emotion: From neural underpinnings and methodological considerations to psychopathology and social cognition},
  journal   = {Psychophysiology},
  year      = {2018},
  volume    = {55},
  number    = {1},
  doi       = {10.1111/psyp.13028}
}

@article{Bangert2006,
  author = {Bangert, Marc and Peschel, Thomas and Schlaug, Gottfried and Rotte, Michael and Drescher, Dieter and Hinrichs, Hermann and Heinze, Hans-Jochen and Altenmueller, Eckart},
  title = {Shared networks for auditory and motor processing in professional pianists: Evidence from fMRI conjunction},
  journal = {NeuroImage},
  volume = {30},
  pages = {917--926},
  year = {2006},
  doi = {10.1016/j.neuroimage.2005.10.044}
}

@article{Brown2019Acting,
  author = {Brown, Steven and Cockett, Peter and Yuan, Ye},
  title = {The neuroscience of Romeo and Juliet: an fMRI study of acting},
  journal = {Royal Society Open Science},
  volume = {6},
  number = {3},
  pages = {181908},
  year = {2019},
  doi = {10.1098/rsos.181908}
}

@manual{clubSandwich,
    title = {clubSandwich: Cluster-Robust (Sandwich) Variance Estimators with Small-Sample Corrections},
    author = {James E. Pustejovsky},
    year = {2025},
    note = {R package version 0.6.1},
    url = {https://CRAN.R-project.org/package=clubSandwich}
}

@misc{Cui2022,
author = {Xu Cui AND Yongrong Wu AND Jipeng Wu AND Zhiyu You AND Jianbing Xiahou AND Menglin Ouyang},
title = {A review: Music-emotion recognition and analysis based on EEG signals},
journal = {Frontiers in Neuroinformatics},
volume = {16},
year = {2022},
doi = {10.3389/fninf.2022.997282}
}

@inproceedings{da2014bitalino,
  title={BITalino: A novel hardware framework for physiological computing},
  author={Da Silva, Hugo Pl{\'a}cido and Guerreiro, Jos{\'e} and Louren{\c{c}}o, Andr{\'e} and Fred, Ana and Martins, Ra{\'u}l},
  booktitle={International Conference on Physiological Computing Systems},
  volume={2},
  pages={246--253},
  year={2014},
  organization={SciTePress}
}

@incollection{daly2018affective,
  title={Affective brain--computer interfacing and methods for affective state detection},
  author={Daly, Ian},
  booktitle={Brain--Computer Interfaces Handbook},
  pages={147--164},
  year={2018},
  publisher={CRC Press}
}

@article{Davidson1990,
  title={Approach-withdrawal and cerebral asymmetry: Emotional expression and brain physiology: I},
  author={Davidson, Richard J and Ekman, Paul and Saron, Clifford D and Senulis, Joseph A and Friesen, Wallace V},
  journal={Journal of Personality and Social Psychology},
  volume={58},
  number={2},
  pages={330--341},
  year={1990},
  publisher={APA}
}

@article{ehrlich2019closedloop, 
    title = {A closed-loop, music-based brain-computer interface for emotion mediation}, 
    author = {Ehrlich, Stefan K. and Agres, Kat R. and Guan, Cuntai and Cheng, Gernot}, 
    journal = {PLOS ONE}, 
    volume = {14}, 
    number = {3}, 
    pages = {e0213516}, 
    year = {2019}, 
    doi = {10.1371/journal.pone.0213516}
}

@article{Ekman01051992,
    author = {Paul Ekman},
    title = {An argument for basic emotions},
    journal = {Cognition and Emotion},
    volume = {6},
    number = {3-4},
    pages = {169--200},
    year = {1992},
    publisher = {Routledge},
    doi = {10.1080/02699939208411068}
}

@article{Gagnon2003,
    author = {Lise Gagnon and Isabelle Peretz},
    title = {Mode and tempo relative contributions to ``happy-sad'' judgements in equitone melodies},
    journal = {Cognition and Emotion},
    volume = {17},
    number = {1},
    pages = {25--40},
    year = {2003},
    publisher = {Routledge},
    doi = {10.1080/02699930302279}
}

@article{gomez2009valence,
  title={Valence lasts longer than arousal: Persistence of induced moods as assessed by psychophysiological measures},
  author={Gomez, Patrick and Zimmermann, PG and Guttormsen Sch{\"a}r, Sissel and Danuser, Brigitta},
  journal={Journal of Psychophysiology},
  volume={23},
  number={1},
  pages={7--17},
  year={2009},
  publisher={Hogrefe \& Huber Publishers}
}

@ARTICLE{Hildt2021,
AUTHOR={Hildt, Elisabeth},
TITLE={Affective Brain-Computer Music Interfaces—Drivers and Implications},
JOURNAL={Frontiers in Human Neuroscience},
VOLUME={15},
YEAR={2021},
DOI={10.3389/fnhum.2021.711407}
}

@inproceedings{hopkins2023stringesthesia,
  title={Stringesthesia: Dynamically shifting musical agency between audience and performer based on trust in an interactive and improvised performance},
  author={Hopkins, Torin and Doherty, Emily and Ofer, Netta and Weng, Suibi Che-Chuan and Gyory, Peter and Tobin, Chad and Hirshfield, Leanne and Do, Ellen Yi-Luen},
  booktitle={Proceedings of the 18th International Audio Mostly Conference},
  pages={9--16},
  year={2023}
}

@incollection{Juslin2010,
  author    = {Juslin, Patrik N. and Liljestr{\"o}m, Simon and V{\"a}stfj{\"a}ll, Daniel and Lundqvist, Lars-Olov},
  title     = {How does music evoke emotions? Exploring the underlying mechanisms},
  booktitle = {Handbook of Music and Emotion: Theory, Research, Applications},
  editor    = {Juslin, Patrik N. and Sloboda, John A.},
  pages     = {605--642},
  publisher = {Oxford University Press},
  year      = {2010}
}

@book{Juslin2019,
    author = {Juslin, Patrik N.},
    title = {Musical Emotions Explained: Unlocking the Secrets of Musical Affect},
    publisher = {Oxford University Press},
    year = {2019},
    doi = {10.1093/oso/9780198753421.001.0001}
}

@Inbook{juslin2010expression,
  title={Expression and communication of emotion in music performance},
  author={Juslin, Patrik N and Timmers, Renee},
  bookTitle={Handbook of music and emotion: Theory, research, applications},
  publisher={Oxford University Press},
  pages={453--489},
  year={2010}
}

@article{Koelsch2014,
  author    = {Koelsch, Stefan},
  title     = {Brain correlates of music-evoked emotions},
  journal   = {Nature Reviews Neuroscience},
  year      = {2014},
  volume    = {15},
  number    = {3},
  pages     = {170--180},
  doi       = {10.1038/nrn3666}
}

@article{Koerner2017,
author = {Tess K. Koerner AND Yang Zhang},
title = {Application of Linear Mixed-Effects Models in Human Neuroscience Research},
journal = {Brain Sciences},
volume = {7},
issue = {3},
pages = {26},
year = {2017},
doi = {10.3390/brainsci7030026}
}

@article{KollerSchlaud2020,
  author    = {Koller-Schlaud, Kerstin and Querbach, Julia and Behr, Johannes and Str{\"o}hle, Andreas and Rentzsch, Johannes},
  title     = {Test-Retest Reliability of Frontal and Parietal Alpha Asymmetry during Presentation of Emotional Face Stimuli in Healthy Subjects},
  journal   = {Neuropsychobiology},
  year      = {2020},
  volume    = {79},
  pages     = {428--436},
  doi       = {10.1159/000505783}
}

@article{kothe2025lsl,
    author = {Kothe, Christian and Shirazi, Seyed Yahya and Stenner, Tristan and Medine, David and Boulay, Chadwick and Grivich, Matthew I. and Artoni, Fiorenzo and Mullen, Tim and Delorme, Arnaud and Makeig, Scott},
    title = {The lab streaming layer for synchronized multimodal recording},
    journal = {Imaging Neuroscience},
    volume = {3},
    pages = {IMAG.a.136},
    year = {2025},
    doi = {10.1162/IMAG.a.136}
}

@book{hermann2011sonification,
  title={The sonification handbook},
  author={Hermann, Thomas and Hunt, Andy and Neuhoff, John G and others},
  volume={1},
  year={2011},
  publisher={Logos Verlag Berlin}
}

@article{Lin2010,
author = {Yuan-Pin Lin AND Chi-Hong Wang AND Tzyy-Ping Jung AND Tien-Lin Wu AND Shyh-Kang Jeng AND Jeng-Ren Duann AND Jyh-Horng Chen},
title = {EEG-Based Emotion Recognition in Music Listening},
journal = {IEEE Transactions on Biomedical Engineering},
volume = {57},
issue = {7},
pages = {1798-1806},
year = {2010},
doi = {10.1109/tbme.2010.2048568}
}

@article{lme4,
    title = {Fitting Linear Mixed-Effects Models Using {lme4}},
    author = {Douglas Bates and Martin M{\"a}chler and Ben Bolker and Steve Walker},
    journal = {Journal of Statistical Software},
    year = {2015},
    volume = {67},
    number = {1},
    pages = {1--48},
    doi = {10.18637/jss.v067.i01}
}

@article{lmerTest,
    title={lmerTest Package: Tests in Linear Mixed Effects Models},
    volume={82},
    doi={10.18637/jss.v082.i13},
    number={13},
    journal={Journal of Statistical Software},
    author={Kuznetsova, Alexandra and Brockhoff, Per B. and Christensen, Rune H. B.},
    year={2017},
    pages={1--26}
}

@article{Luo2025,
  author    = {Luo, Yuxi and Tang, Min and Fan, Xiaodong},
  title     = {Meta analysis of resting frontal alpha asymmetry as a biomarker of depression},
  journal   = {npj Mental Health Research},
  year      = {2025},
  volume    = {4},
  doi       = {10.1038/s44184-025-00117-x}
}

@article{Mennella2017,
  author    = {Mennella, Raffaella and Patron, Elisabetta and Palomba, Daniela},
  title     = {Frontal alpha asymmetry neurofeedback for the reduction of negative affect and anxiety},
  journal   = {Behaviour Research and Therapy},
  year      = {2017},
  volume    = {92},
  pages     = {32--40},
  doi       = {10.1016/j.brat.2017.02.002}
}

@book{miranda2014guide,
  title={Guide to brain-computer music interfacing},
  author={Miranda, Eduardo Reck and Castet, Julien},
  year={2014},
  publisher={Springer}
}

@article{nakagawa2013general,
  title={A general and simple method for obtaining R2 from generalized linear mixed-effects models},
  author={Nakagawa, Shinichi and Schielzeth, Holger},
  journal={Methods in ecology and evolution},
  volume={4},
  number={2},
  pages={133--142},
  year={2013},
  publisher={Wiley Online Library}
}

@incollection{Ochsner2014,
  author    = {Ochsner, Kevin N. and Gross, James J.},
  title     = {The neural bases of emotion and emotion regulation: A valuation perspective},
  editor    = {Gross, James J.},
  booktitle = {Handbook of Emotion Regulation},
  edition   = {2nd},
  pages     = {23--42},
  year      = {2014},
  publisher = {Guilford Press}
}

@article{Pfurtscheller1999,
  title={Event-related EEG/MEG synchronization and desynchronization: Basic principles},
  author={Pfurtscheller, Gert and Lopes da Silva, Fernando H},
  journal={Clinical Neurophysiology},
  volume={110},
  number={11},
  pages={1842--1857},
  year={1999},
  publisher={Elsevier}
}

@article{Ramirez2015,
author = {Rafael Ram{\'i}rez AND Manuel Palencia-Lefler AND Sergio Giraldo AND Zacharias Vamvakousis},
title = {Musical neurofeedback for treating depression in elderly people},
journal = {Frontiers in Neuroscience},
volume = {9},
year = {2015},
doi = {10.3389/fnins.2015.00354}
}

@article{Ramirez2020,
author = {Rafael Ram{\'i}rez AND Sergio Giraldo AND Zacharias Vamvakousis},
title = {Brain-Computer Music Interface for Music Expression},
journal = {array. the journal of the ICMA},
pages = {86-88},
year = {2020},
doi = {10.25370/array.v20152530}
}

@article{Reznik2018,
  author    = {Reznik, Samantha J. and Allen, John J. B.},
  title     = {Frontal asymmetry as a mediator and moderator of emotion: An updated review},
  journal   = {Psychophysiology},
  year      = {2018},
  volume    = {55},
  number    = {1},
  doi       = {10.1111/psyp.12965}
}

@article{Ribeiro2019,
  author    = {Ribeiro, Fabiana Silva and Santos, Flávia Heloísa and Albuquerque, Pedro Barbas and Oliveira-Silva, Patrícia},
  title     = {Emotional Induction Through Music: Measuring Cardiac and Electrodermal Responses of Emotional States and Their Persistence},
  journal   = {Frontiers in Psychology},
  year      = {2019},
  volume    = {10},
  pages     = {451},
  doi       = {10.3389/fpsyg.2019.00451}
}

@article{ruiz2020panas,
    title={PANAS Internacional Revisado: Propiedades psicom{\'e}tricas en una muestra internacional latina},
    author={Ruiz-P{\'e}rez, Jos{\'e} Ignacio and Melo-Gonz{\'a}lez, Vanessa and Velandia-Amaya, Sergio Nicol{\'a}s and Rodr{\'\i}guez-Mesa, Luz Stella and Monroy, C{\'e}sar Alfonso Vel{\'a}zquez},
    journal={Universitas Psychologica},
    volume={19},
    pages={1--11},
    year={2020},
    doi={10.11144/Javeriana.upsy19.pirp}
}

@article{Russell1980,
author = {James A. Russell},
title = {A circumplex model of affect.},
journal = {Journal of Personality and Social Psychology},
volume = {39},
issue = {6},
pages = {1161-1178},
year = {1980},
doi = {10.1037/h0077714}
}

@misc{Sabu2022,
author = {Priya Sabu AND Ivo V. Stuldreher AND Daisuke Kaneko AND Anne-Marie Brouwer},
title = {A Review on the Role of Affective Stimuli in Event-Related Frontal Alpha Asymmetry},
journal = {Frontiers in Computer Science},
volume = {4},
year = {2022},
doi = {10.3389/fcomp.2022.869123}
}

@article{Sitaram2017,
  title={Closed-loop brain training: the science of neurofeedback},
  author={Sitaram, Ranganatha and Ros, Tomas and Stoeckel, Luke and Haller, Sven and Scharnowski, Frank and Lewis-Peacock, Jarrod and Weiskopf, Nikolaus and Blefari, Maria Laura and Rana, Mohit and Oblak, Ethan and others},
  journal={Nature Reviews Neuroscience},
  volume={18},
  number={2},
  pages={86--100},
  year={2017}
}

@article{SokolHessner2022,
  author = {Sokol-Hessner, Peter and Wing-Davey, Mark and Illingworth, Scott and Fleming, Stephen M. and Phelps, Elizabeth A.},
  title = {The actor's insight: Actors have comparable interoception but better metacognition than non-actors},
  journal = {Emotion},
  volume = {22},
  number = {7},
  pages = {1544--1553},
  year = {2022},
  doi = {10.1037/emo0001080}
}

@article{supper2014sublime,
  title={Sublime frequencies: The construction of sublime listening experiences in the sonification of scientific data},
  author={Supper, Alexandra},
  journal={Social Studies of Science},
  volume={44},
  number={1},
  pages={34--58},
  year={2014},
  publisher={Sage Publications Sage UK: London, England}
}

@article{Takabatake2021,
author = {Kazuhiko Takabatake AND Naoto Kunii AND Hirofumi Nakatomi AND Seijiro Shimada AND Kei Yanai AND Megumi Takasago AND Nobuhito Saito},
title = {Musical Auditory Alpha Wave Neurofeedback: Validation and Cognitive Perspectives},
journal = {Applied Psychophysiology and Biofeedback},
volume = {46},
issue = {4},
pages = {323-334},
year = {2021},
doi = {10.1007/s10484-021-09507-1}
}

@article{VanDerVinne2017,
  author    = {Van Der Vinne, Vincent and Vollebregt, Madelon A. and Van Putten, Michel J. A. M. and Arns, Martijn},
  title     = {Frontal alpha asymmetry as a diagnostic marker in depression: Fact or fiction? A meta-analysis},
  journal   = {NeuroImage: Clinical},
  year      = {2017},
  volume    = {16},
  pages     = {79--87},
  doi       = {10.1016/j.nicl.2017.07.006}
}

@inbook{vickers2016aesthetics,
author = {Vickers, Paul and Hogg, Bennett and Worrall, David},
year = {2017},
month = {04},
pages = {89-109},
title = {Aesthetics of sonification: Taking the subject-position},
isbn = {9781472485403},
publisher = {Routledge},
doi = {10.4324/9781315569628-6}
}

@article{Welch1967,
author = {P. Welch},
title = {The use of fast Fourier transform for the estimation of power spectra},
journal = {IEEE Transactions on Audio and Electroacoustics},
volume = {15},
issue = {2},
pages = {70-73},
year = {1967},
doi = {10.1109/tau.1967.1161901}
}

@incollection{williams2018bci,
  title={BCI for music making: then, now, and next},
  author={Williams, Duncan AH and Miranda, Eduardo R},
  booktitle={Brain--Computer Interfaces Handbook},
  pages={193--206},
  year={2018},
  publisher={CRC Press}
}

@article{Wu2023,
author = {Dongrui Wu AND Bao-Liang Lu AND Bin Hu AND Zhigang Zeng},
title = {Affective Brain–Computer Interfaces (aBCIs): A Tutorial},
journal = {Proceedings of the IEEE},
volume = {111},
issue = {10},
pages = {1314-1332},
year = {2023},
doi = {10.1109/jproc.2023.3277471}
}

@article{Yu2021,
author = {Zhaoxia Yu AND Michele Guindani AND Steven F. Grieco AND Lujia Chen AND Todd C. Holmes AND Xiangmin Xu},
title = {Beyond t test and ANOVA: applications of mixed-effects models for more rigorous statistical analysis in neuroscience research},
journal = {Neuron},
volume = {110},
issue = {1},
pages = {21-35},
year = {2021},
doi = {10.1016/j.neuron.2021.10.030}
}

@article{Zatorre2007,
  author = {Zatorre, Robert J. and Chen, Joyce L. and Penhune, Virginia B.},
  title = {When the brain plays music: auditory--motor interactions in music perception and production},
  journal = {Nature Reviews Neuroscience},
  volume = {8},
  number = {7},
  pages = {547--558},
  year = {2007},
  doi = {10.1038/nrn2152}
}

@article{Zotev2016AmygdalaFAA,
  title={Correlation between amygdala BOLD activity and frontal EEG asymmetry during real-time fMRI neurofeedback training in depression},
  author={Zotev, Vadim and others},
  journal={NeuroImage: Clinical},
  volume={11},
  pages={224--238},
  year={2016},
  doi={10.1016/j.nicl.2016.02.003}
}

\end{sloppy}
\end{document}